# Deep Convolutional Neural Networks as Generic Feature Extractors

Lars Hertel[*][†], Erhardt Barth[†], Thomas Käster[†‡] and Thomas Martinetz[†]
[*]Institute for Signal Processing, University of Luebeck, Germany
Email: hertel@isip.uni-luebeck.de
[†]Institute for Neuro- and Bioinformatics, University of Luebeck, Germany
Email: {barth, kaester, martinetz}@inb.uni-luebeck.de
[‡]Pattern Recognition Company GmbH, Luebeck, Germany

*Abstract*—Recognizing objects in natural images is an intricate problem involving multiple conflicting objectives. Deep convolutional neural networks, trained on large datasets, achieve convincing results and are currently the state-of-the-art approach for this task. However, the long time needed to train such deep networks is a major drawback. We tackled this problem by reusing a previously trained network. For this purpose, we first trained a deep convolutional network on the ILSVRC-12 dataset. We then maintained the learned convolution kernels and only retrained the classification part on different datasets. Using this approach, we achieved an accuracy of $67.68\,\%$ on CIFAR-100, compared to the previous state-of-the-art result of $65.43\,\%$. Furthermore, our findings indicate that convolutional networks are able to learn generic feature extractors that can be used for different tasks.

## I. Introduction

Recognizing objects in natural images is an intricate task for a machine, involving multiple conflicting objectives. The effortlessness of the human brain deceives the complex underlying process. Inspired by the mammalian visual system, convolutional neural networks were proposed [1]–[4]. They are the state-of-the-art approach for various pattern recognition tasks. Unlike many other learning algorithms, convolutional networks combine both *feature extraction* and *classification*. The advantage of this approach was impressively demonstrated by LeCun et al. [4] on MNIST and Krizhevsky et al. [5] on ILSVRC-12, achieving better results than previous learning methods.

A schematic representation of a convolutional network is shown in Figure 1. The given network comprises five different layers, i.e. input, convolution, pooling, fully-connected and output layer. The input layer specifies a fixed size for the input images, i.e. images may have to be resized accordingly. The image is then convolved with multiple learned kernels using shared weights. Next, the pooling layer reduces the size of the image while trying to maintain the contained information. These two layers compose the feature extraction part. Afterwards, the extracted features are weighted and combined in the fully-connected layer. This represents the classification part of the convolutional network. Finally, there exists one output neuron for each object category in the output layer.

Recent results indicate that very deep networks achieve even better results on various benchmarks [6], [7]. Moreover,

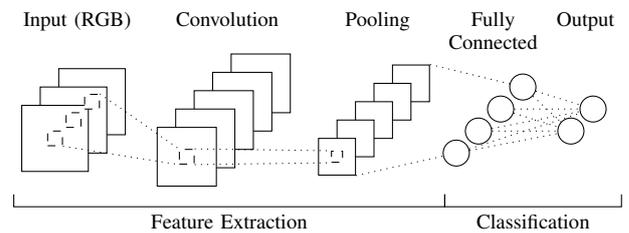

Fig. 1. Schematic diagram of a convolutional neural network. The network comprises five different layers. Both feature extraction and classification are learned during training.

an ensemble of multiple networks and additional training data are often used to further increase the performance [8], [9]. Thus, the general formula for a convincing performance are seemingly multiple deep convolutional networks with many layers and a huge amount of training data.

One drawback of this trend, however, is the long time needed to train such deep networks. To tackle this problem, we reused a previously trained network. For this purpose, we first trained a convolutional network on a large dataset, maintained the learned feature extraction part, and only *retrained* the classification part on multiple different datasets. We then compared the results to a *full* training, i.e. both feature extraction and classification, of the same network on the same dataset.

## II. Datasets

In this work, we used four different datasets, namely ILSVRC-12, MNIST, CIFAR-10 and CIFAR-100. We used ILSVRC-12 to *pretrain* our network and MNIST, CIFAR-10 and CIFAR-100 to *retrain* our network afterwards. In the following section, we will briefly introduce each dataset. An overview of some statistics of all four datasets is given in Table I.

The IMAGENET [10] dataset contains more than 14 million labeled high-resolution color images of natural objects and scenes belonging to over 21,000 different categories. The images were collected from the web and labeled by humans using *Amazon Mechanical Turk*[1]. A subset of these images is taken as an annual competition called the *ImageNet*

---
[1] http://www.mturk.com

TABLE I. STATISTICS OF THE USED DATASETS.

| Dataset | Classes | Dimension | | | No. Images | | |
|---|---|---|---|---|---|---|---|
| | | Width | Height | Depth | Training | Validation | Test |
| MNIST | 10 | 28 | 28 | 1 | 60,000 | - | 10,000 |
| CIFAR-10 | 10 | 32 | 32 | 3 | 50,000 | - | 10,000 |
| CIFAR-100 | 100 | 32 | 32 | 3 | 50,000 | - | 10,000 |
| ILSVRC-12 | 1,000 | - | - | 3 | 1,281,167 | 50,000 | 150,000 |

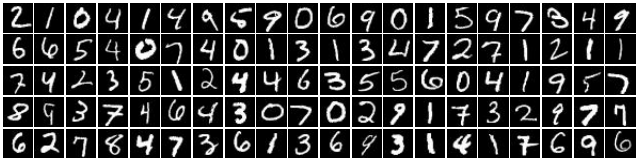

Fig. 2. 100 randomly selected images from MNIST. The dataset represents handwritten digits. It has ten different classes, one for each digit from zero to nine. The digits are centered and normalized in size. MNIST comprises 70,000 grayscale images with dimensions of $28 \times 28$.

*Large-Scale Visual Recognition Challenge* (ILSVRC). We used the dataset from the competition in 2012 (ILSVRC-12). It consists of nearly 1.5 million color images, which are varying in size, and 1,000 different categories.

The MNIST [11] dataset contains grayscale images of handwritten digits. 100 randomly selected images from MNIST are shown in Figure 2. It possesses ten different categories, namely one for each digit from zero to nine. Each grayscale image has a fixed size of $28 \times 28$ pixels. The digits are centered inside the image and normalized in size. In total, MNIST contains 70,000 images, split into 60,000 training and 10,000 test images.

The CIFAR-10 and CIFAR-100 [12] datasets contain small color images of natural objects. An excerpt of 96 randomly chosen images is shown in Figure 3. They are labeled subsets of the *80 Million Tiny Images*[2] database. CIFAR-10 possesses ten and CIFAR-100 possesses 100 different categories, respectively. Each color image has a fixed size of $32 \times 32$ pixels. In total, they both consist of 60,000 images, split into 50,000 training and 10,000 test images.

## III. METHODS

The architecture of our trained convolutional network is shown in Table II. It is based on the architecture proposed by Krizhevsky et al. [5]. The network comprises 24 layers. In particular, five convolution and three maximum pooling layers with different square kernel sizes and kernel strides. Moreover, we added zero padding in some cases to obtain convenient sizes of the feature maps. As a nonlinear activation function, we settled for the rectified linear unit [13]. Moreover, layers of dropout [14] were applied after each fully-connected layer. The probability to randomly drop a unit in the network is $50\,\%$. Finally, to obtain a probability distribution, we employed a softmax layer.

Just like Krizhevsky et al. [5], we trained the network on ILSVRC-12 and reached comparable results, i.e. a top-1 accuracy of $59.23\,\%$ and a top-5 accuracy of $83.58\,\%$. For this purpose, we resized all images of the dataset to a fixed size of $256 \times 256$ pixels and randomly cropped a subimage of $227 \times 227$ pixels. This increased the number of training images by a factor of 900. We used the deep learning framework *Caffe* [15] to train our convolutional networks. It allowed us to employ the GPU[3] of our computer for faster training. This resulted in a speedup of approximately ten, compared to training with the CPU[4].

[2]http://groups.csail.mit.edu/vision/tinyimages/

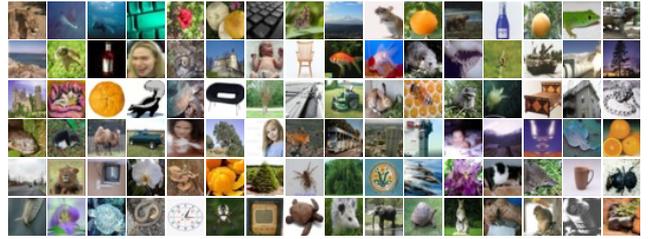

Fig. 3. 96 randomly selected images from CIFAR-100. The dataset represents 100 different natural objects. It contains 60.000 color images with dimensions of $32 \times 32$. CIFAR-10 possesses the same statistics, except that it has ten classes with 6,000 images each.

96 learned kernels of the first convolution layer are shown in Figure 4. Each kernel is a color image of $11 \times 11$ pixels. Though the colors are difficult to interpret, Gabor-like filters of different frequencies and orientations can be recognized. They are used to extract edges of the input image. Further learned kernels of deeper convolution layers with a size of $5 \times 5$ and $3 \times 3$ are too small to provide any noticeable information and are therefore not shown.

We then maintained the feature extraction part – i.e. the learned convolution kernels – of the previously trained network and only retrained the classification part – i.e. the fully-connected layers – on different datasets, namely MNIST, CIFAR-10 and CIFAR-100. To adjust the three datasets to our network architecture, we had to alter two things. First, to resize the images to $256 \times 256$ pixels. Note that we had to convert the grayscale images of MNIST to color images by simply copying the grayscale channel three times. Secondly, to change the number of output neurons to the number of different object categories, depending on the dataset.

To evaluate the performance of the trained network on an independent test set, we averaged ten different predictions of a single image as proposed by Krizhevsky et al. [5]. For this purpose, we averaged the output of the four corner crops and the center crop of the input image and – except for MNIST – additionally mirrored each image along the vertical axis. We then trained the same network a second time for each dataset. This time, however, we trained the full network, i.e. both feature extraction and classification, and compared both obtained results.

The networks were trained for 30 epochs. An epoch means a complete training cycle over all images of the training set. We started with a fixed base learning rate $\eta = 0.001$ and decreased it by a factor of ten to $\eta = 0.0001$ after 20 epochs. Furthermore, we selected a momentum value of $\mu = 0.9$, a weight decay value of $\lambda = 0.0005$, and a batch size of $\beta = 80$. These parameters were determined based on a validation set.

---

[3]NVIDIA GeForce GTX 770 with 2 GB of memory
[4]Intel Core i7-4770

TABLE II. ARCHITECTURE OF OUR IMPLEMENTED CONVOLUTIONAL NETWORK.

| No. | Layer | Dimension | | | Kernel | Stride | Padding |
|---|---|---|---|---|---|---|---|
| | | Width | Height | Depth | | | |
| 0 | Input | 227 | 227 | 3 | - | - | - |
| 1 | Convolution | 55 | 55 | 96 | 11 | 4 | - |
| 2 | Relu | 55 | 55 | 96 | - | - | - |
| 3 | Pooling | 27 | 27 | 96 | 3 | 2 | - |
| 4 | Normalization | 27 | 27 | 96 | - | - | - |
| 5 | Convolution | 27 | 27 | 256 | 5 | 1 | 2 |
| 6 | Relu | 27 | 27 | 256 | - | - | - |
| 7 | Pooling | 13 | 13 | 256 | 3 | 2 | - |
| 8 | Normalization | 13 | 13 | 256 | - | - | - |
| 9 | Convolution | 13 | 13 | 384 | 3 | 1 | 1 |
| 10 | Relu | 13 | 13 | 384 | - | - | - |
| 11 | Convolution | 13 | 13 | 384 | 3 | 1 | 1 |
| 12 | Relu | 13 | 13 | 384 | - | - | - |
| 13 | Convolution | 13 | 13 | 256 | 3 | 1 | 1 |
| 14 | Relu | 13 | 13 | 256 | - | - | - |
| 15 | Pooling | 6 | 6 | 256 | 3 | 2 | - |
| 16 | Fully Connected | 1 | 1 | 4096 | - | - | - |
| 17 | Relu | 1 | 1 | 4096 | - | - | - |
| 18 | Dropout | 1 | 1 | 4096 | - | - | - |
| 19 | Fully Connected | 1 | 1 | 4096 | - | - | - |
| 20 | Relu | 1 | 1 | 4096 | - | - | - |
| 21 | Dropout | 1 | 1 | 4096 | - | - | - |
| 22 | Fully Connected | 1 | 1 | 1000 | - | - | - |
| 23 | Softmax | 1 | 1 | 1000 | - | - | - |

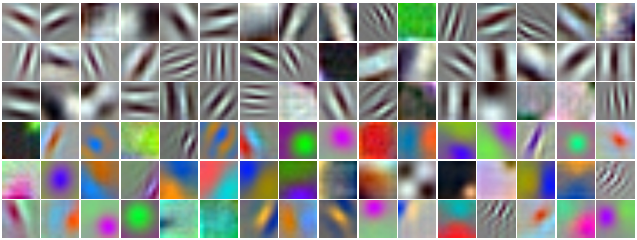

Fig. 4. 96 learned kernels of the first convolution layer. Each kernel has dimensions of $11 \times 11 \times 3$. They were maintained when retraining a network.

## IV. RESULTS

Our results are shown in Figure 5. The horizontal axis represents the number of epochs. The vertical axis shows the accuracy of the independent test set. Note that, for simplicity, the results in Figure 5 were computed only for the center crop of the input images. The solid lines represent full training, i.e. the learning of both feature extraction and classification. The dashed lines represent the retraining of the classification part only. The red lines correspond to MNIST, the blue lines correspond to CIFAR-10 and the green lines correspond to CIFAR-100, respectively.

The results show that, regarding full training and retraining, comparable accuracy rates are achieved after 30 epochs for all three datasets. More precisely, while both rates for MNIST are nearly identical, the rate for retraining is slightly worse in case of CIFAR-10 and slightly better in case of CIFAR-100 compared to full training.

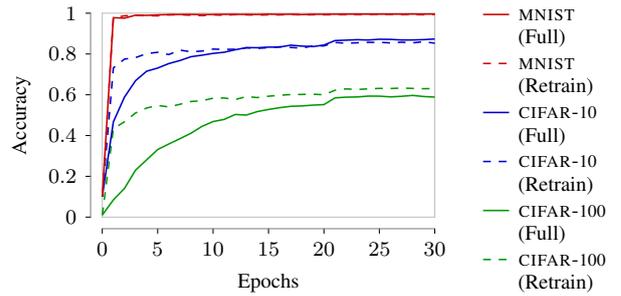

Fig. 5. Comparison between full training and retraining on three different datasets. We trained the network for 30 epochs. The solid lines represent full training, i.e. both feature extraction and classification. The dashed lines represent retraining, i.e. classification only.

Note that training the full network is considerably slower than retraining: it takes about ten epochs to achieve the accuracy that the retrained network has after a single epoch in case of CIFAR-100 and about eight epochs in case of CIFAR-10, respectively. Further note the slight increase in accuracy for CIFAR-10 and CIFAR-100 after 20 epochs due to the decrease of the learning rate.

Further results are given in Table III. It shows the accuracy of the independent test set after 30 epochs both for full training and retraining. Moreover, the state-of-the-art results from the literature are given as reference values. Note that this time the accuracy rates were calculated by averaging the predictions of multiple crops of the input image. For CIFAR-10 and CIFAR-100 we averaged the outputs of ten crops. Since we did not mirror the images for MNIST, for obvious reasons, we only averaged the outputs of five crops.

The results additionally underline the comparable accuracy rates between full training and retraining for the respective dataset. As for the CIFAR-100 dataset, retraining the network achieves an accuracy rate of 67.68 %, which is 2.25 % better than the state-of-the-art accuracy.

## V. DISCUSSION

Deep convolutional neural networks are the state-of-the-art approach for object recognition. One drawback, however, is the long time needed to train such deep networks, especially on large datasets. We tackled this problem by reusing the feature extraction part of a previously trained network and only retrained the classification part on multiple different datasets.

As expected, our findings show that this approach considerably reduces the necessary amount of training time. Our results indicate a speedup by a factor of up to ten, depending on the dataset. Besides the three presented datasets, we applied the proposed method to a self-made dataset with nearly 500,000 images from Flickr belonging to 110 different object categories and obtained very good results in a short time. We did not present the results, however, since the accuracy rates of the fully trained network are not available for comparison.

TABLE III. RESULTS OF THE TEST SET FOR FULL TRAINING AND RETRAINING AFTER 30 EPOCHS.

| Dataset | Accuracy (%) | | |
|---|---|---|---|
| | Full Training | Retraining | State of the Art |
| MNIST | 99.68 | 99.54 | 99.79 [22] |
| CIFAR-10 | 89.99 | 89.14 | 91.78 [23] |
| CIFAR-100 | 63.65 | **67.68** | 65.43 [23] |

More interestingly, both fully trained and retrained networks achieve comparable accuracy rates on all three datasets. The learned feature extractor from our pretrained network is therefore applicable to multiple situations. Even though the feature extractor was trained on ILSVRC-12, containing natural images and scenes, it still achieves excellent results even on digits from the MNIST dataset. This finding indicates that further datasets can be classified with the same feature extractor.

Our experiments confirm and extend the results reported by Razavian et al. [16], Donahue et al. [17] and Girshick et al. [18], who have also trained linear and nonlinear classifiers on features obtained from deep learning with convolutional networks.

Note that we trained all networks – including the one from pretraining – in a supervised manner using backpropagation [19]. The obtained feature extractor was therefore trained for a specific purpose. Its generic characteristics are somewhat surprising. Previously, mostly unsupervised algorithms, like sparse coding and related representation learning algorithms, have been used for pretraining. This approach achieves state-of-the-art results on the STL-10 [20] dataset, as shown by Miclut et al. [21].

However, our results show that it is also possible to perform supervised pretraining and obtain excellent results. This approach even improved the accuracy on CIFAR-100, compared to a fully trained deep convolutional network. This suggests that more appropriate kernels were learned from ILSVRC-12 than from CIFAR-100 itself.

## VI. CONCLUSIONS

Reusing a previously trained convolutional network not only vastly reduces the necessary time for training, but also achieves comparable results regarding the full training of the network. For particular datasets, the accuracy is even increased. This finding is especially relevant for practical applications, e.g. when only limited computing power or time is available. Our results indicate the existence of a generic feature extractor concerning the three used datasets. To either support or reject this hypothesis, further research for multiple datasets in different situations should be considered.